\documentclass[aps,prfluids,preprint,groupedaddress]{revtex4-2}
\usepackage{graphicx}
\usepackage{hyperref}
\usepackage{xcolor}
    \definecolor{blue2}{rgb}{0, 0.4470, 0.7410}
    \definecolor{orange}{rgb}{0.8500, 0.3250, 0.0980}
    \definecolor{yellow1}{rgb}{0.9290, 0.6940, 0.1250}
    \definecolor{yellow}{rgb}{0.9922,0.9059,0.1451}
    \definecolor{purple}{rgb}{0.2667, 0.0039, 0.3294}
    \definecolor{green}{rgb}{0.1294, 0.5686,0.5490}

\usepackage{amsmath}
\usepackage{booktabs}
\begin{document}

\newcommand{\mutext}{µ}


\title{DefocusTrackerAI - A Generalized Framework for the Automatic Detection of Defocused Particle Images}


\author{Gonçalo Coutinho}
\email[]{goncalo.coutinho@tecnico.ulisboa.pt}
\affiliation{IN+ Center for Innovation, Technology and Policy Research, Instituto Superior Técnico, University of Lisbon, Lisbon, Portugal}


\author{Ana S. Moita}
\affiliation{IN+ Center for Innovation, Technology and Policy Research, Instituto Superior Técnico, University of Lisbon, Lisbon, Portugal}
\affiliation{CINAMIL - Military Academy Research Center, Militart Academy, Portugal}

\author{António L. N. Moreira}
\affiliation{IN+ Center for Innovation, Technology and Policy Research, Instituto Superior Técnico, University of Lisbon, Lisbon,Portugal}

\author{Massimiliano Rossi}
\affiliation{Department of Industrial Engineering, Alma Mater Studiorum University of Bologna, Bologna, Italy}

\date{\today}

\begin{abstract}
The present work introduces \textit{DefocusTrackerAI}, a generalized deep-learning framework 
for the automatic detection and position estimation of defocused particle images from any kind of optical configuration without compromising uncertainty and recall, intended as a follow-up of the open-source project \textit{DefocusTracker}. We selected the deep neural network architecture from the direct comparison of two well-known object detection models, Faster R-CNN and YOLOv9, trained on a diverse and feature-rich synthetic image set containing astigmatic and non-astigmatic defocused particle images of varying diameters. The model evaluation on synthetic data showed that, first, YOLOv9 outperforms Faster R-CNN, achieving higher recall and lower uncertainty, particularly at high particle image densities; and second, that YOLOv9 provides enhanced spatial resolution, with uncertainty values between 0.1 and 0.4 pixels for particle image densities ($N_s$) up to 0.5, outperforming state-of-the-art algorithms. We demonstrated that our models are able to detect astigmatic and non-astigmatic defocused particle images in multiple optical setups with varying lighting conditions. In addition, we successfully applied our models on real DPT experiments, including fluorescence and shadowgraph data, showing that they can be used beyond conventional DPT applications, including the tracking of sprays and droplets. A pre-trained, ready-to-use version of \textit{DefocusTrackerAI} based on YOLOv9 is available at 
\href{https://gitlab.com/goncalo.coutinho/defocustrackerAI-main/-/tree/7e0f11f649ebad50e20dca5b9545f26ca303ebe0/}{DefocusTrackerAI GitLab} and can be used for automatic detection of defocused particle images of any kind with high accuracy. In combination with a suitable calibration approach for the depth position, it can be used as an effective first step for three-dimensional defocusing particle tracking.

\end{abstract}


\maketitle
\section{Introduction}\label{sect:Introduction}
The growth of the artificial intelligence field of machine learning has continuously led to advancements in computer vision tasks requiring object detection. Industries such as autonomous driving \cite{autonomousdriving}, surveillance systems \cite{surveillance}, and aerial monitoring \cite{aerial} have benefited from this growth, and measurement science technologies are no exception. In recent years, we have seen a growing number of studies adopting different deep neural network (DNN) architectures for the detection of defocused particle images. Defocusing particle tracking (DPT) is a single-camera measurement technique that allows to track particles in three-dimensional space \cite{Barnkob2015general}. For optical setups that provide particle images that vary solely with the out-of-plane coordinate ($z$), the method takes advantage of the depth-dependent shape of the defocused particle images to map the depth position. The literature comprises a list of different mapping methods, including model functions \cite{Cierpka2010, Rossi2014optimization, Fuchs2016situ}, normalized cross-correlation \cite{Barnkob2020, Rossi2020GDPT}, and DNNs \cite{Konig2020, Barnkob2021, Dreisbach2022}. Briefly, model functions based methods utilize a parametric model to characterize the particle image shape as a function of depth. Normalized cross-correlation compares the detected particle images with a reference template to assess similarity. The DNN approaches follow a two-stage framework, where the first network identifies particle positions, and the second estimates their depth.

One of the main concerns within the DPT community has been the development of novel methodologies that enhance the detection of defocused particle images by minimizing uncertainty in the measured coordinates ($x, y, z$) and improving particle detection in densely populated images, thereby improving accuracy and spatial resolution. Increasing the number of detected particles reduces the required number of recorded images and, consequently, the overall measurement time. However, this is not seen as a trivial task. As previously observed \cite{Barnkob2021}, higher particle detection rates often come at the expense of increased measurement uncertainty. This trade-off was recently overcome with DNNs \cite{Ratz2023}. In particular, a Faster Region-Based Convolutional Neural Network (Faster R-CNN) \cite{Ren2015}, trained with astigmatic defocused particle images, demonstrated a significant increase in particle detection while maintaining low measurement uncertainty.

Motivated by these developments, we believe that recent advances in computer vision with single stage object detectors such as the You Only Look Once (YOLO) v9 architecture may enhance the detection of defocused particle images \cite{wang2024yolov9}. Compared to the Faster R-CNN \cite{Ren2015}, YOLOv9 features a lightweight architecture while achieving state-of-the-art detection performance on the widely recognized MS COCO dataset \cite{COCOdataset}. Despite its promising potential, this architecture has yet to be evaluated for the detection of defocused particle images. 

\begin{figure*}[tph!]
\centerline{\includegraphics[width=\linewidth]{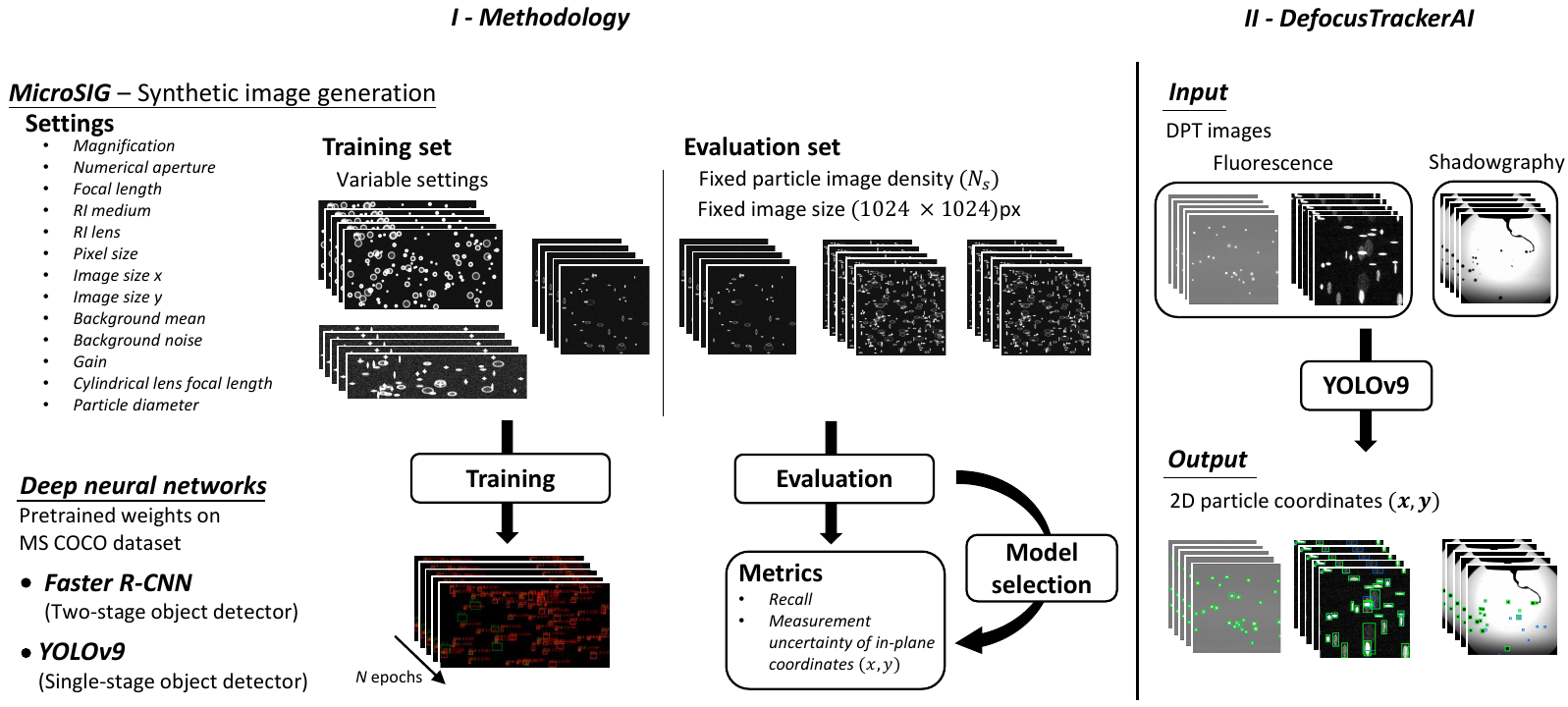}}
\caption{I - Flowchart representing the methodology used to generate the synthetic image sets for training and evaluating of the object detection models using MicroSIG\,\cite{Rossi2020MicroSig}, and the selection of the object detection model, including the training and evaluation. II - Flowchart representing the DefocusTrackerAI framework, including the input (DPT images - fluorescence or shadowgraphy), the inference using YOLOv9 and the output (2D particle coordinates $(x,y)$). }
\label{fig:Flowchart}
\end{figure*}

In another perspective, DNN approaches have the major drawback of needing to be trained on the specific type of defocused particle images provided by the experimental setup \cite{Dreisbach2022, Franchini2020cut}. In turn, this makes their practical application not straightforward, especially for non-experienced users. While this is most likely inevitable for out-of-plane mapping (i.e., $z$), it would be possible to train a DNN model on a large variety of astigmatic and non-astigmatic defocused particle images, ensuring generalization to a wide range of optical configurations, with varying lighting conditions. Although limited to a two-dimensional space (2D), it would leave room for other research groups to take advantage of the detection capabilities and cascade, if needed, another framework to associate the detections to the out-of-plane coordinate ($z$). For instance, one could cascade a second DNN \cite{Sax2022}, or choose between a model function \cite{Cierpka2010} and normalized cross-correlation \cite{Rossi2020GDPT}. 
	
In light of the previous discussion, the present work proposes a generalized deep-learning framework (Fig.\,\ref{fig:Flowchart}: II - \textit{DefocusTrackerAI}) that is ready to use and can be applied directly for the detection of astigmatic and non-astigmatic defocused particle images from any kind of optical setup, and is intended as a follow-up to the open-source \textit{DefocusTracker} project \cite{Barnkob2021defocustracker}. To that end, first, we perform the selection of DNN model based on the comparison between two well-known object detection models, Faster R-CNN and YOLOv9, trained on a diverse and feature-rich synthetic image set containing astigmatic and non-astigmatic defocused particle images of varying diameters. Although experimental images offer additional features not present in synthetic ones, synthetic images can be generated with relatively low effort and provide direct access to ground truth. In contrast, obtaining diverse and feature-rich experimental image set is time-consuming, manual labelling is labour-intensive and may introduce bias.
Then, we evaluate the performance of our models using a second set of synthetic images with defocused particle images with different levels astigmatism and particle image densities to access both the generalization to multiple optical configurations and spatial resolution. In addition, we provide a comparison with previous methods using a reference dataset \cite{Barnkob2021}. Ultimately, the object detection models are tested on experimental images from three DPT measurements, using fluorescence and shadowgraphy, to assess their ability to generalize to different experimental conditions. The remainder of this work is organized as follows: In Sec.\,\ref{sect:Methods}, we describe the preparation of the synthetic datasets and object detection models; performance results on synthetic and experimental images are presented in Sec.\,\ref{sect:SyntImages} and Sec.\,\ref{sect:ExpImages}, respectively; discussion is given in Sec.\,\ref{sect:Discussion} and conclusions are provided in Sec.\,\ref{sec:Conclusions}.

\section{Methods}\label{sect:Methods}
\subsection{Preparation of synthetic datasets}\label{subsec:Methods-PreparationSynt}
The object detection models tested for \textit{DefocusTrackerAI} were trained using synthetic images generated by MicroSIG, a synthetic image generator for astigmatic and non-astigmatic defocused particle images based on ray tracing \cite{Rossi2020MicroSig}, see Fig.\,\ref{fig:Flowchart}. The choice of using synthetic over experimental images is driven by the fact that synthetic images provide direct access to ground-true coordinates and bounding boxes of the defocused particle images, which are crucial for the training, validation, and testing phases of DNN. Furthermore, they allow for precise control of the particle image density ($N_s$) \cite{Barnkob2020}, which favours direct comparisons between different models and algorithms, as we will see in Sec.\,\ref{sect:SyntImages}, Fig.\,\ref{fig:NSCritic}. The particle image density ($N_s$) is given by the ratio between the sum of the areas of the defocused particle images ($A_p$) and the full image area ($A_I$)

\begin{equation}\label{eq:NS}
    N_s = \frac{1}{A_I} \sum_i A^{(i)}_p \approx N_p \frac{\bar{A_p}}{A_I}.
\end{equation}

In contrast, experimental images require manual data annotation, a task that, while feasible, is time-consuming and intensive, particularly given the large datasets necessary for training DNN. Moreover, this procedure introduces potential biases and additional sources of uncertainty, which can ultimately degrade the performance of the DNN \cite{Dreisbach2022}.

\begin{figure}[tph!]
\centerline{\includegraphics[width=0.65\linewidth]{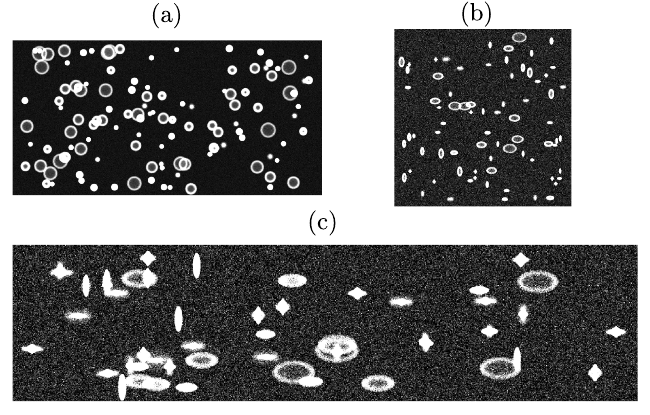}}
\caption{Examples of the synthetic image set used to train the object detection models, including different image size, particle diameter, particle image density, noise level, as well as astigmatic and non-astigmatic defocused particle images (a)-(c). 
}
\label{fig:SyntheticTrainingSet}
\end{figure}

To create a diverse and feature-rich image set for training, we simulated seventeen different DPT experiments by varying the settings of MicroSIG, including numerical aperture, focal length, background noise, particle diameter, and also astigmatism by controlling the focal length of the cylindrical lens, yielding non-astigmatic particle images when set to zero --- Fig.\,\ref{fig:Flowchart}. The choice of these parameters was made based on our experience conducting DPT measurements in different setups and with multiple optical configurations \cite{CoutinhoPRFluids2023, Coutinho2023biascorrection, Coutinho2024}. Additionally, we varied the image size in the range between $256$ and $1024$ pixels, both in length and width, thus the particle image density. In total, we generated $500$ images for each case, combining to a total of 8500 images and 507,500 annotations. An overview of the synthetic image set used for training is given in Fig.\,\ref{fig:SyntheticTrainingSet}(a)-(c).

\begin{figure*}[tph!]
\centerline{\includegraphics[width=\textwidth]{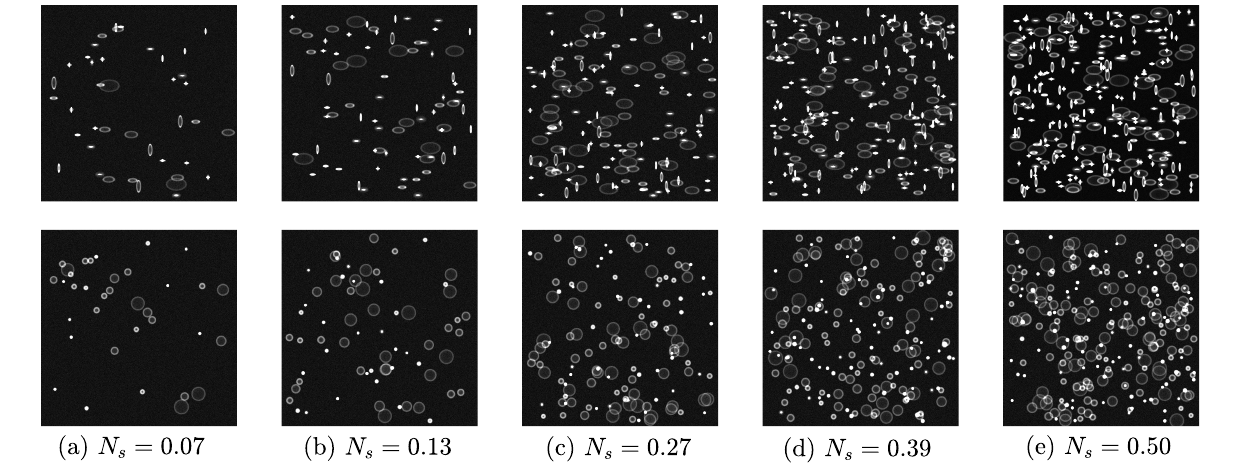}}
\caption{Examples of the synthetic image set used to test the object detection models, astigmatic (top row) and non-astigmatic (bottom row) defocused particle images at a minimum particle image size of $16$\,pixels for five different particle image densities ($N_s$) (a)-(e). 
}
\label{fig:SyntheticTestSet}
\end{figure*}

For the evaluation of the object detection models, 
we generated a different set of synthetic data ($1024\times1024$-pixel images) with different levels astigmatism to access the generalization to multiple optical configurations, and more importantly, with specific particle image densities, $N_s = [0.03,\,0.07,\,0.13,\,0.20,\,0.27,\,0.39,\,0.50]$, to access spatial resolution. Examples of the synthetic image set used for model evaluation are given in Fig.\,\ref{fig:SyntheticTestSet}(a)-(e). These data was further split into cases with minimum particle image size ($d_\text{min}$) of either $8$ or $16$ pixels. These values were chosen based on two key factors: first, our previous DPT experiments typically involved a minimum particle image diameter of approximately 16 pixels (e.g., \cite{Coutinho2024}); and second, DNN often struggle to detect small objects, particularly those around $8-10$ pixels in size, in densely populated images. In total, we created 28 test sets, each case containing a total of 100 frames.

\subsection{Object detection models}\label{subsec:Methods-Faster}

In this work, we compare two object detection models, the first being the two-stage object detector Faster R-CNN \cite{Ren2015}, which recently led to state-of-the-art results in the DPT domain \cite{Ratz2023}. The Faster R-CNN employs a residual network with 50 layers (ResNet50) \cite{he2016} as its backbone, along with a feature pyramid network (FPN) \cite{lin2017}. Briefly, the ResNet50 backbone is responsible for extracting features from input images, while the FPN enhances this process by constructing a multi-scale feature pyramid, enabling the detector to leverage both low- and high-resolution features. As a two-stage object detector, it uses a Region Proposal Network (RPN) to process the feature maps and generate a set of candidate object regions by predicting their bounding boxes and a score for each box. In the second stage, the region proposals are further refined and classified. The object centre is then obtained from the output bounding box, and as a post-processing step, the detections with scores below a given confidence score are removed, as we will see later in Sec.\,\ref{sect:SyntImages}.

To train the Faster R-CNN model, we leveraged transfer learning and used the pre-trained weights from the MS COCO dataset \cite{COCOdataset}, available through TorchVision's pre-trained models in PyTorch, thereby reducing training time, required amount of training data, and risk of overfitting --- Fig.\,\ref{fig:Flowchart}. The model was trained using the stochastic gradient descent optimizer (SGD) with a momentum of 0.9, a batch size of four and a $L_2$ regularization of $1e^{-4}$. We used a learning rate scheduler, including three warm-up epochs, in which the learning rate increased from $3e^{-5}$ to $1e^{-4}$, followed by a cosine annealing decay with the number of epochs. To reduce the risk of overfitting and improve generalization, we used data augmentation techniques, including image translation and scaling, as well as brightness, contrast, saturation adjustment, and inverted colour with different probabilities. An overview of these techniques is shown later in Fig.\,\ref{fig:DataAug}. Additionally, we implemented an early stop with a patience of four, i.e., the training is terminated if there is no improvement after four epochs.

On the other side of object detection models, we have single-stage object detectors, such as the YOLO series \cite{redmon2016}. Unlike two-stage detectors such as Faster R-CNN, which rely on an RPN, YOLO directly predicts bounding box coordinates, object confidence scores, and class probabilities in a single forward pass. Furthermore, it benefits from lighter architecture, lower computational cost, and lower inference times compared to the Faster R-CNN. We used YOLOv9 version \cite{wang2024yolov9}, which introduces the generalized efficient layer aggregation network (GELAN) as the backbone and programmable gradient information (PGI) to improve detection performance and computational efficiency. GELAN is an evolution of the efficient layer aggregation network (ELAN) \cite{wang2023yolov7}, and is designed to enhance the ELAN capabilities to extract multilevel features from the input image, while maintaining inference speed. Recent comparisons showed that feature maps from a 50-layer GELAN outperform those produced by ResNet50 \cite{wang2024yolov9}. The PGI is a mechanism designed to optimize gradient propagation throughout the network. In general, this architecture leads to improved model convergence and overall accuracy, as demonstrated by recent state-of-the-art results on the COCO dataset \cite{wang2024yolov9}.

We considered two different variations of the YOLOv9 architecture: YOLOv9-m and YOLOv9-c. The network was trained using the SGD optimizer with a momentum of 0.937, batch size of four, $L_2$ regularization of $1e^{-4}$. Again, we used data augmentation to reduce the risk of overfitting (Fig.\,\ref{fig:DataAug}). Following \cite{wang2024yolov9}, we used three warm-up epochs to update the bias only, set the initial learning rate to $1e^{-4}$, and used a linear learning rate decay. The models were trained on Google Colab with an NVIDIA L4 GPU with $22.5$\,GB of RAM. The reader can find the settings for the training and data augmentation in Appendix A, Tab. \ref{table:trainingsettings} and \ref{Table:Augmentation}.

\section{Analysis of synthetic images}\label{sect:SyntImages}

\begin{figure*}[tph!]
	\centerline{\includegraphics[width=\textwidth]{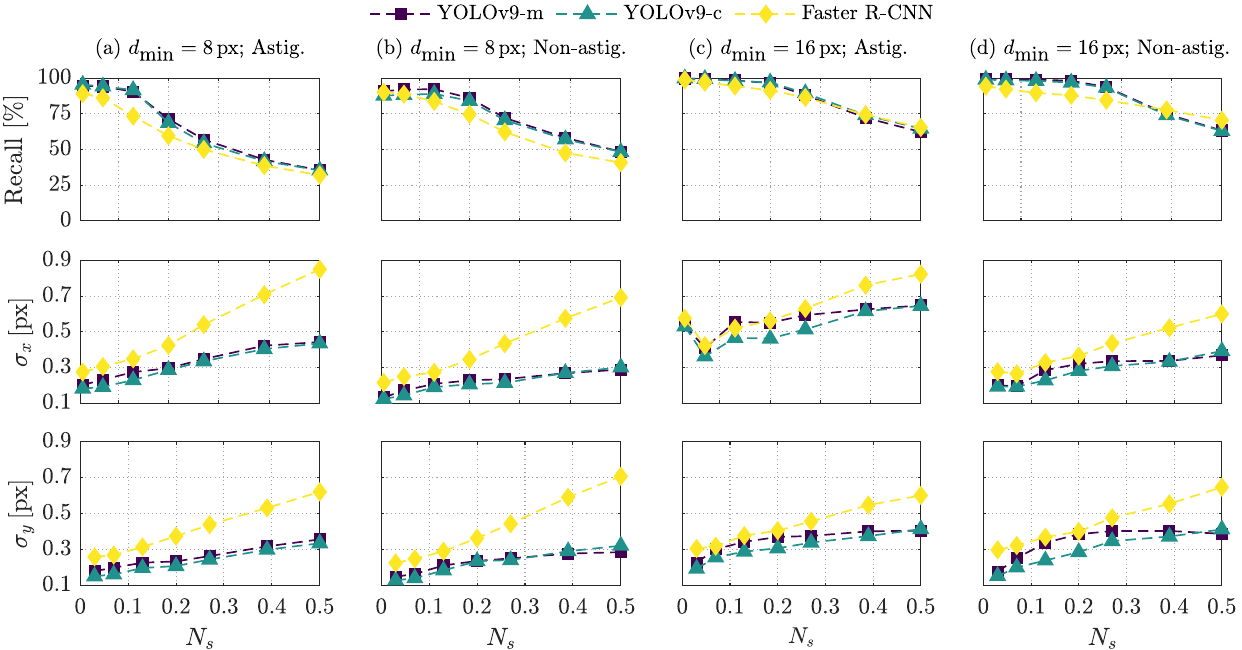}}
	\caption{Recall and uncertainty in the in-plane coordinates ($\sigma_{x,y}$) as function of the particle image density ($N_s$) for the synthetic test sets using the Faster R-CNN and YOLOv9 models at a confidence score of 0.8. (a) $d_\text{min}=8$\,px, Astig. (b) $d_\text{min}=8$\,px, Non-astig. (c) $d_\text{min}=16$\,px, Astig. (d) $d_\text{min}=16$\,px, Non-astig. }
	\label{fig:PerformanceSyntheticTestSet}
\end{figure*}

The models were evaluated using the synthetic test sets from Fig.\,\ref{fig:SyntheticTestSet} with a confidence score of 0.8, which represents the probability that the detected object is a defocused particle image. Following previous works, we measured the performance by means of recall and measurement uncertainty in the in-plane coordinates ($\sigma_{x,y}$). Recall is defined as the ratio of true positive (TP) detections to the total number of true positive and false negative (FN) detections: $\textrm{Recall} = \frac{\textrm{TP}}{\textrm{TP}+\textrm{FN}}$. In other words, recall quantifies the detection rate by measuring the proportion of objects that were correctly identified as defocused particle images. Uncertainty is defined as the root mean square of the error between the measured coordinates ($x_i$,\, $y_i$) and ground-true coordinates ($x_0$,\,$y_0$): $\sigma_x = \sqrt{\frac{\sum^N_{i=1} (x_0 - x_i)^2}{N}}$, where $N$ is the number of true positive detections \cite{Barnkob2020, Barnkob2021}. The recall and uncertainty are shown both in Fig.\,\ref{fig:PerformanceSyntheticTestSet} for the astigmatic and non-astigmatic cases, considering minimum particle image diameters ($d_\text{min}$) of $8$ and $16$ pixels. 

In general, recall (first row) indicates that for smaller defocused particle images ($d_\text{min}=8$\,px), recall sits around $90\,\%$ for low particle image densities ($N_s < 0.13$) and has a steep decrease as $N_s$ increases. Specifically, recall drops close to $66\,\%$ at $N_s=0.2$, and further drops to $32-45\,\%$ at $N_s=0.5$. This decrease is a natural consequence of the growing number of particles per frame and the larger occurrence of overlapping particles. For defocused particle images with larger diameters ($d_\text{min}=16$\,px), recall shows an improvement close to $7$\,\% at low particle image densities ($N_s < 0.13$), over $10\,\%$ at $N_s = 0.2$, and $20\,\%$ at $N_s=0.27$. Among the tested models, YOLOv9-m and YOLOv9-c achieve the highest recall rates, reaching $96\,\%$ and $89\,\%$ at $N_s=0.2$ and $N_s=0.27$, respectively. Faster R-CNN, while slightly less effective, follows with recall values of 87\,\% and 84\,\% for the same particle image densities. 

\begin{figure}[tph!]
\centerline{\includegraphics[width=0.5\linewidth]{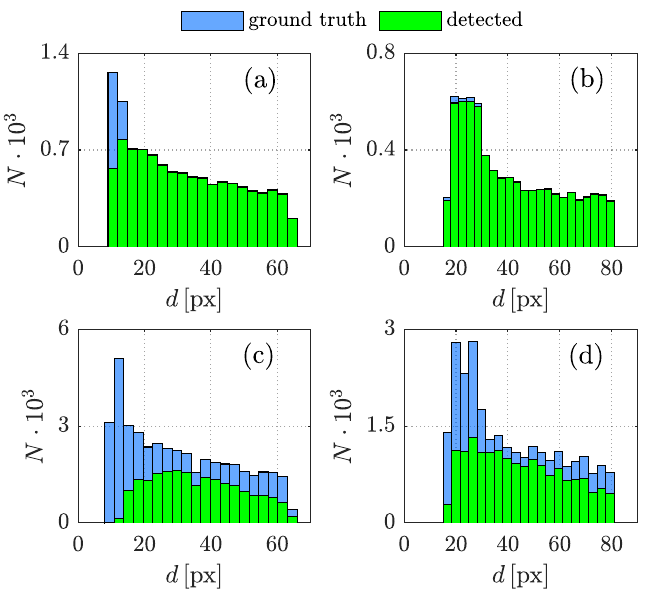}}
\caption{Particle image size distribution ($d$) between the ground truth and the true positive detections performed by YOLOv9-m for the non-astigmatic particle images. (a) $d_{min}=8$\,pixel, $N_s=0.13$. (b) $d_{min}=16$\,pixel, $N_s=0.13$. (c) $d_{min}=8$\,pixels, $N_s=0.5$. (d) $d_{min}=16$\,pixels, $N_s=0.5$.}
\label{fig:sizeDist}
\end{figure}

\begin{figure}[tph!]
\centerline{\includegraphics[width=0.85\linewidth]{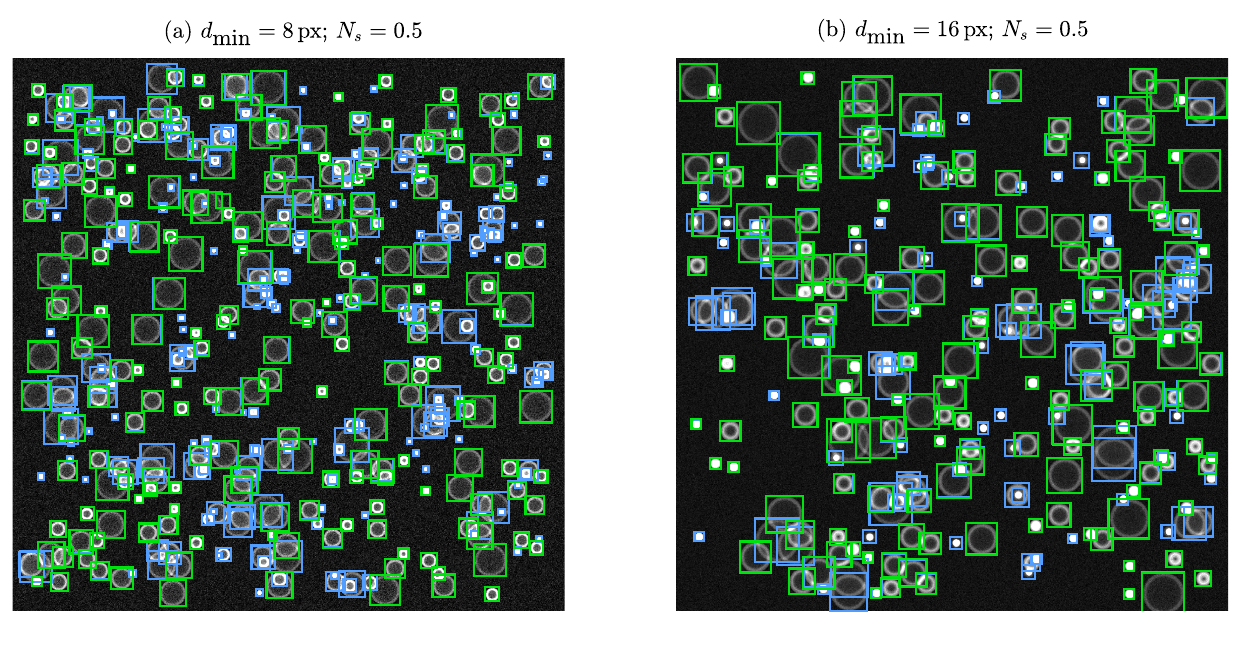}}
\caption{Ground truth (blue box) and detected particle images (green box) for the non-astigmatic particle images in $1024\times1024$-pixel image, with $d_{min}=[8,\,16]$\,pixels at $N_s=0.5$ (a)-(b). }
\label{fig:sizeDist_Image}
\end{figure}

The lower performance on defocused particle images with smaller diameter is a well-known challenge in object detection tasks, particularly in densely crowded images. To further examine this limitation, we compare in Fig.\,\ref{fig:sizeDist} the distribution of the particle image diameter ($d$) between ground truth and true positive detections performed by YOLOv9-m for non-astigmatic defocused particle images at $N_s = 0.13$ and $N_s=0.5$, considering both $d_{min}$ equal to $8$ and $16$\,pixels. From Fig.\,\ref{fig:sizeDist}(a), we observe that the inferior performance is mainly due to the limited capacity of the model to detect particle images within the $8$ to $14$ pixel range. In contrast, the test case with larger defocused particles (i.e., $d>14$\,px) shows improved performance (Fig.\,\ref{fig:sizeDist}(b)). As $N_s$ increases to 0.5, the limitation of the model to detect small particle images becomes even more pronounced, with barely no detections in the 8 to 14 pixel range, see Fig.\,\ref{fig:sizeDist}(c). It comes with no surprise that with increasing occurrence of overlapping particles, detecting small defocused particle images becomes even more complicated. Such a result is particularly evident in Fig.\,\ref{fig:sizeDist}(c) and \ref{fig:sizeDist}(d). To aid further comparisons, Fig.\,\ref{fig:sizeDist_Image} shows the ground truth (green box) and the true positive detections (red box) in synthetic images at $N_s=0.5$,  for both $d_{min}=[8,\,16]$\,pixels.

Regarding the measurement uncertainty in the in-plane coordinates ($\sigma_{x,y}$), our results show that the centre of the defocused particle images is determined with subpixel accuracy --- Fig.\,\ref{fig:PerformanceSyntheticTestSet} (second and third row). Without surprise, the uncertainty increases with $N_s$. As already discussed, the growing number of particles makes detection challenging, thereby determining the bounding box and the particle image centre see, e.g., Fig.\,\ref{fig:sizeDist_Image}(a). Interestingly, the YOLOv9 models outperform Faster R-CNN in terms of uncertainty, particularly in densely crowded images, where the difference in the uncertainty ranges between 0.2 and 0.4 pixels, see Fig.\,\ref{fig:PerformanceSyntheticTestSet}. Moreover, while YOLOv9-c exhibits slightly lower uncertainty than YOLOv9-m, this marginal improvement may not justify its use, given its increased complexity, higher number of parameters and higher computational cost \cite{wang2024yolov9}. This argument is supported by the comparative analysis in Table\,\ref{tab:yolo_models}, which evaluates the performance of the YOLOv9 models in terms of number of parameters, processing speed (i.e., frames per second (FPS)) and computational cost (i.e., giga floating point operations per second (GFLOPs)) on a synthetic image set (Non-astigmatic; $d_\textrm{min}=16$\,px; $N_s=0.5$).

\begin{table}[tph!]
\centering
\small
\setlength{\tabcolsep}{10pt} 
 {\fontsize{11}{11}\selectfont 
\begin{tabular}{@{}lccc@{}}
\toprule
Model & FPS & GFLOPs & Params (M) \\ \midrule
YOLOv9-m & 33 & 167.28 & 32.55  \\
YOLOv9-c & 21 & 302.88 & 50.70  \\ \bottomrule
\end{tabular}
}
\caption{Performance metrics of YOLOv9 models on synthetic image set (Non-astigmatic; $d_\textrm{min}=16$\,pxs; $N_s=0.5$), including frames per second (FPS), giga floating point operations per second (GFLOPs) and number of parameters. Obtained on Google Colab with an NVIDIA L4 GPU with $22.5$\,GB of RAM.}
\label{tab:yolo_models}
\end{table}

The greater uncertainty in the $x$ coordinate (Fig.\,\ref{fig:PerformanceSyntheticTestSet}(c) --- $d_\text{min}=16$\,px; Astig.), obtained for astigmatic particle images may arise from an imbalanced training set. We believe that augmenting the training set will reduce the uncertainty to levels comparable to those observed in the other cases.

\begin{figure*}[tph!]
	\centerline{\includegraphics[width=\textwidth]{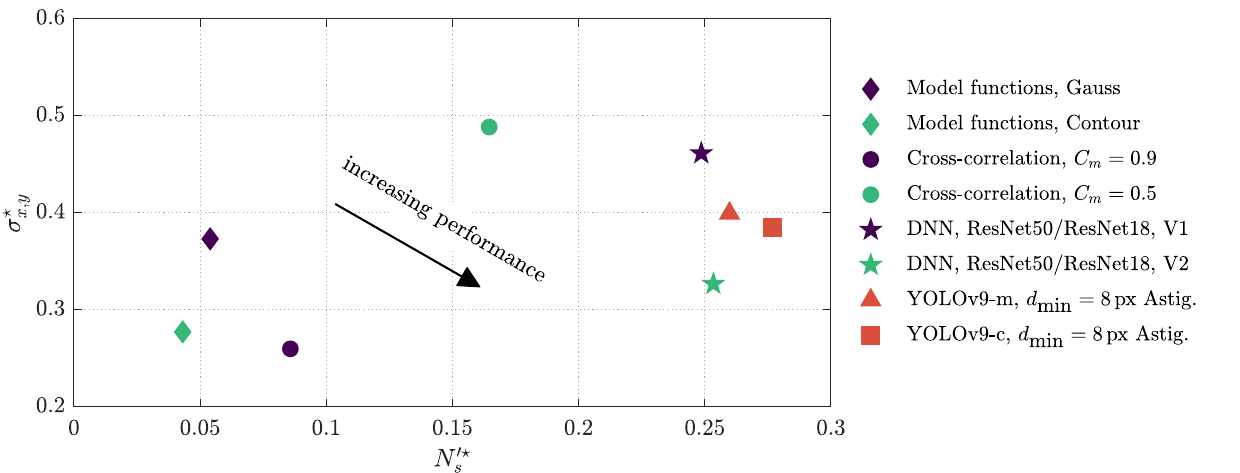}}
	\caption{Comparison between our YOLOv9-m and YOLOv9-c models with previous methods, including the model functions and cross-correlation from \cite{Barnkob2021}, and DNN from \cite{Ratz2023}, with respect to the maximum measured particle image density ($N^{\prime \star}_s$) and the corresponding uncertainty ($\sigma^\star_{x,y}$).}
	\label{fig:NSCritic}
\end{figure*}

Compared to previous methods (e.g., \cite{Ratz2023}), our models, YOLOv9-m and YOLOv9-c, establish new benchmarks for recall in the DPT domain, particularly for $N_s<0.2$ --- ranges commonly encountered in DPT experiments \cite{Coutinho2024, leister2021flow, sachs2023particle}. Moreover, this is achieved without increasing the measurement uncertainty ($\sigma^\star_{x,y}$), since, in fact, our uncertainty is comparable to the Faster R-CNN model presented by Ratz et al.\,\cite{Ratz2023}. Following previous works \cite{Barnkob2021, Ratz2023}, further comparisons between YOLOv9-m and YOLOv9-c models with model-based functions and normalized cross-correlation \cite{Barnkob2021}, and DNN \cite{Ratz2023} are provided in Fig.\,\ref{fig:NSCritic} by evaluating the maximum measured particle image density ($N^{\prime \star}_s$) and the corresponding uncertainty ($\sigma^\star_{x,y}$). The comparison is performed with respect to the reference dataset proposed in \cite{Barnkob2021} based on astigmatic defocused particle images, that is equivalent to our case presented in Fig.\,\ref{fig:PerformanceSyntheticTestSet}(c) ($d_\text{min}=8$\,px, Astigmatic). The measured particle image density ($N^{\prime}_s$) is determined similarly to $N_s$ (Eq.\,(\ref{eq:NS})), yet with the particle image area $A_p$ of the true positive detections. Here, $\sigma^\star_{x,y}$ represents the median of the uncertainties ($\sigma_x,\,\sigma_y$) from Fig.\,\ref{fig:PerformanceSyntheticTestSet}. Fig.\,\ref{fig:NSCritic} shows that the YOLOv9-m and YOLOv9-c models enhance the detection of defocused particle images with a higher measured particle image density ($N^{\prime \star}_s$) (higher spatial resolution) and comparable uncertainty. However, unlike previous works, which were specifically trained either for astigmatic \cite{Barnkob2021, Konig2020, Ratz2023} or non-astigmatic defocused particle images \cite{Dreisbach2022, Sax2022}, our models were trained to handle both types effectively, across multiple optical configurations with different lighting conditions. Their generalization to different types of defocused particle image makes them an attractive solution for less experienced users with limited knowledge of DPT, eliminating the need for a steep initial learning curve. 

\section{Analysis of experimental images}\label{sect:ExpImages}

\subsection{Detection performance}

\begin{figure*}[tph!]
	\centerline{\includegraphics[width=\textwidth]{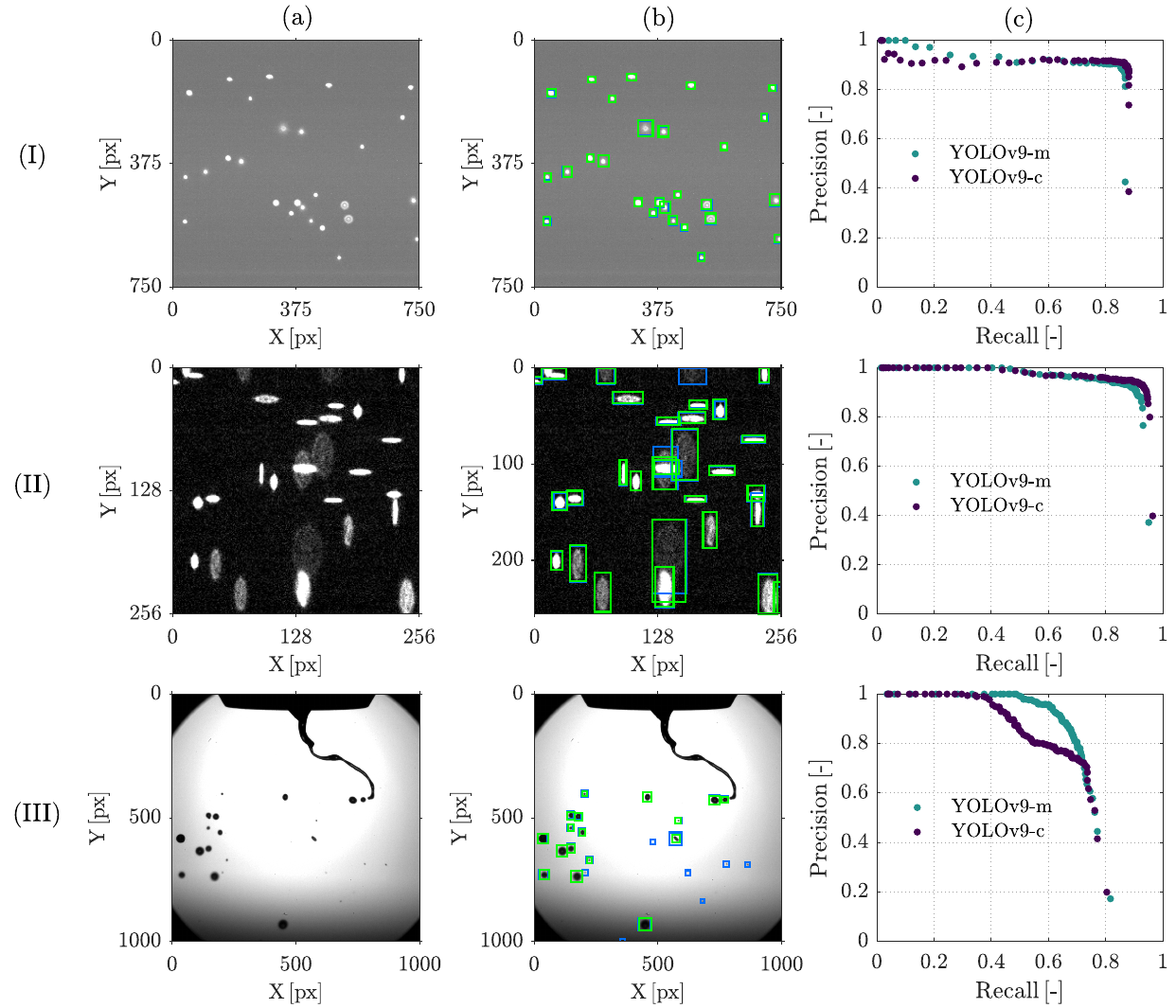}}
	\caption{Performance of the YOLOv9-m and YOLOv9-c models on experimental data, including two recordings using fluorescence (I \cite{Barnkob2021}; II \cite{coutinhoUChannel}), and one using shadowgraphy (III \cite{InesSprayLXLASER}). (a) Raw images. (b) Ground truth (blue box) and corresponding detections (green box) obtained with YOLOv9-m at a confidence score of 0.8. (c) Detection performance by means of precision and recall.}
	\label{fig:ExpImages_DPT}
\end{figure*}

With the benchmark performance of the YOLOv9-m and YOLOv9-c models established on the synthetic test sets, we now assess their performance on experimental data from three distinct DPT measurements, two using fluorescence and one using shadowgraphy. The fluorescence data include a test set from DefocusTracker \cite{Barnkob2021defocustracker} with non-astigmatic particle images (I - $1280\times750$-pixel images), and a measurement in a 180-degree curved artery model with astigmatic particle images (II - $728\times256$-pixel images) \cite{coutinhoUChannel}. Regarding shadowgraphy, we used raw images from a liquid jet breakup of an air-assisted atomizer (III - $1024\times1024$-pixel images) \cite{InesSprayLXLASER}. Each case was manually annotated to obtain a total of 500 labelled defocused particle images. As noted by \cite{Dreisbach2022}, manually labelled images can be used to assess detection performance in terms of precision and recall but not accuracy, due to inherent bias and uncertainty introduced by manual annotation. Fig.\,\ref{fig:ExpImages_DPT} shows the different data (I, II and III), (a) including raw images, (b) raw images with ground truth (green boxes) and corresponding detections (red boxes) for a confidence score of 0.8, and ultimately (c) the detection performance by means of precision and recall. Precision is defined as the ratio of true positive detections to the total number of true positive and false positive (FP) detections: $\textrm{Precision} = \frac{\textrm{TP}}{\textrm{TP}+\textrm{FP}}$. For purposes of plotting, the experimental images were cut to a square size. 

Fig.\,\ref{fig:ExpImages_DPT} shows the generalization of YOLOv9-m and YOLOv9-c to different types of defocused particle images, in experimental recordings using fluorescence. Specifically, the models reliably identify defocused particle images of varying sizes, including both astigmatic and non-astigmatic cases, as evidenced by the precision-recall curves for datasets I and II. Thereby being a robust and versatile solution for the automatic detection of defocused particle images in varying DPT experiments. Regarding shadowgraphy, while, in general, the models detected droplets originating from the breakup jet, their performance is lower. In this type of experiments, the non-controllable droplet size presents a challenge, as smaller droplets may appear (6-8 pixel range), making detection more difficult. This is evident in Fig.\,\ref{fig:ExpImages_DPT}\,(III), where a few small droplets on the right side of the image remained undetected by the models. In addition, given that both models were trained and tested under identical conditions, we are currently unable to attribute the performance drop of YOLOv9-c to a specific factor. Nevertheless, these data highlight the ability of our models to generalize as well to different experimental conditions, e.g., varying contrast and illumination, typically associated with shadowgraph measurements. More importantly, it shows the potential of our models to be applied beyond the typical DPT domain, including the tracking of sprays and droplets.

\subsection{Influence of data augmentation}\label{sec:DataAug}

Data augmentation techniques were used to introduce greater variability into the synthetic training set, thereby enhancing robustness against diverse imaging conditions, noise levels, particle size, and distributions. In particular, we applied geometric transformations, such as translation and scaling, as well as colour-based augmentations, as described in Table\,\ref{Table:Augmentation}. We did not consider the mosaic augmentation often used with YOLOv9 \cite{wang2024yolov9}, since it is better suited for multi-object detection tasks and offers limited benefits for our single-object detection tasks in rather uniform-background scenarios as is the case of DPT measurements.

\begin{figure*}[tph!]
	\centerline{\includegraphics[width=1\textwidth]{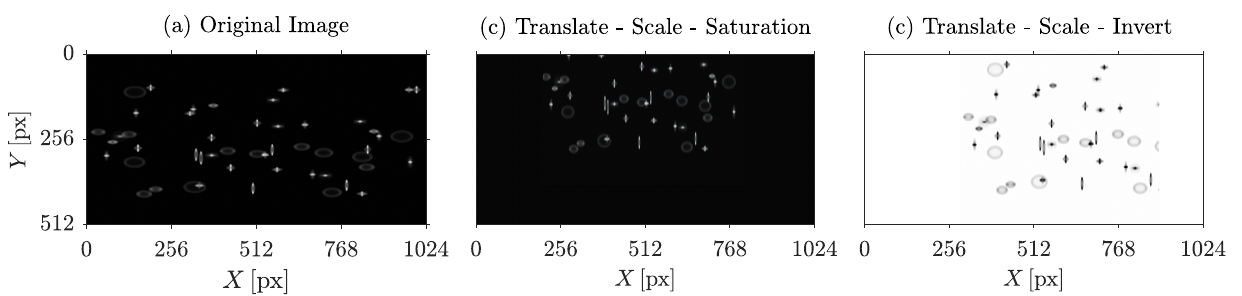}}
	\caption{Overview of the data augmentation techniques. (a) Original image. (b) Image translated, stretched and with decreased brightness. (c) Image translated, stretched and with colour inversion.}
	\label{fig:DataAug}
\end{figure*}

\begin{figure}[tph!]
\centerline{\includegraphics[width=\linewidth]{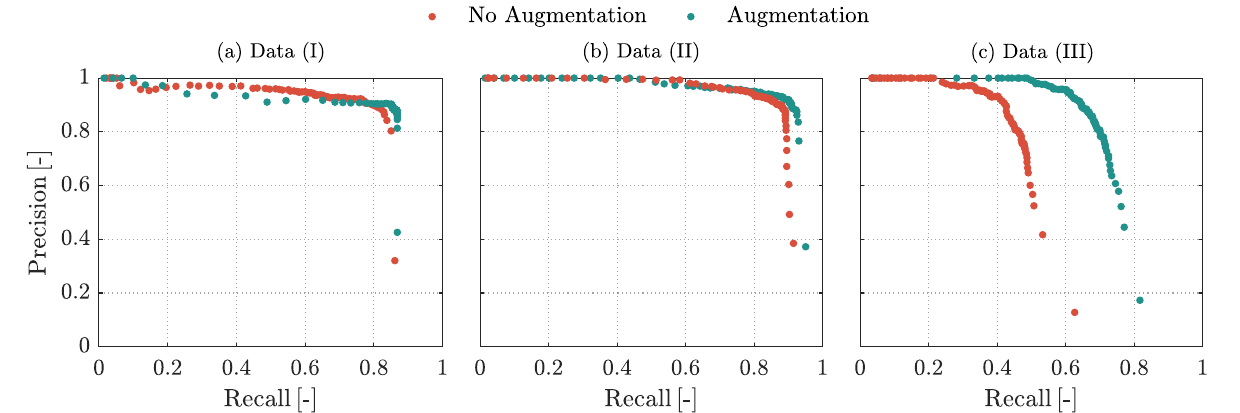}}
\caption{Influence of data augmentation on the detection performance of YOLOv9-m by means of precision and recall, including the three DPT recordings, using fluorescence (I \cite{Barnkob2021}; II \cite{coutinhoUChannel}) and shadowgraphy (III \cite{InesSprayLXLASER}) (a)-(c).}
\label{fig:RecallAUg}
\end{figure}

Fig.\,\ref{fig:DataAug} provides an overview of such data augmentation techniques. To quantify the improvement in the performance derived from these augmentation techniques and support future developments with DNNs in the DPT domain, we also trained a baseline model without data augmentation. Given the computational costs associated with training these models, we made the decision to focus solely on YOLOv9-m. The models were again evaluated on the DPT measurements (I, II and III) by means of precision and recall, as shown in Fig.\,\ref{fig:RecallAUg}.

For the data obtained with fluorescence (I and II), the augmentation techniques had a limited effect, leading only to $3$\,\% of improvement. Since the synthetic images were designed to mimic the conditions found in DPT experiments, the model was already well trained without the aid of data augmentation. In contrast, for shadowgraphy, which exhibits a significant variability in the image characteristics, the application of these strategies resulted in an improvement of approximately $30\,\%$. The greater variability introduced in training made the model more capable of treating dynamic and complex environments, not included in the training set.

\section{Discussion}\label{sect:Discussion}

We demonstrated that the YOLOv9 architectures surpass the Faster R-CNN as the preferred model for detecting defocused particle images, with improved recall and considerably lower uncertainty in the in-plane coordinates ($x,y$), especially for higher particle image densities ($N_s>0.2$) (Fig.\,\ref{fig:PerformanceSyntheticTestSet}). More importantly, YOLOv9-m and YOLOv9-c architectures outperformed the state-of-the-art algorithm from \cite{Ratz2023} with enhanced spatial resolution while maintaining uncertainty levels, as shown in Fig.\,\ref{fig:NSCritic} --- Increasing the number of detected particles reduces the required number of recorded images and, consequently, the overall measurement time. In addition, the performance in both astigmatic and non-astigmatic defocused particle images demonstrated its generalization to a wide range of optical configurations, thereby making \textit{DefocusTrackerAI} an attractive solution for less experienced users with limited knowledge of DPT. Up to now, DNN approaches have focused solely on astigmatic \cite{Konig2020, Ratz2023} or non-astigmatic defocused particle images \cite{Dreisbach2022, Sax2022}. 

The validation on real DPT experiments using fluorescence showed that our YOLOv9 models can reliably detect defocused particle images in the experimental data \cite{Barnkob2021defocustracker, coutinhoUChannel}. Concerning the DPT experiments using shadowgraphy, the models yielded a lower performance due to the presence of small droplets ($6$-$8$\,px) --- a well-known challenge in compute vision tasks --- nonetheless showing the potential to be deployed well beyond the typical DPT domain, including the tracking of sprays and droplets \cite{InesSprayLXLASER}. It is yet worth noting that the generalization capabilities to the different lighting conditions found in shadowgraphy was only possible due to data augmentation strategies.

Regarding the training of DNN, our results showed that by using a diverse and feature-rich synthetic image set, one can train object detection models for DPT experiments, ensuring generalization to real DPT data. This represents a significant milestone, since the low cost and effort involved in generating synthetic data using tools like MicroSIG \cite{Rossi2020MicroSig} makes this training approach attractive and accessible to a wide range of DPT applications. In comparison, the generation of training sets with experimental images \cite{Ratz2023}, or even hybrid datasets combining experimental and synthetic images \cite{Dreisbach2022}, are considerably more resource-intensive. However, we believe that the incorporation of experimental images can further improve the performance of object detection models, especially in scenarios with a highly variable background or extreme variations in illumination, e.g., \cite{Dreisbach2022}.

Despite the advances on the detection of defocused particle images, our models struggled to identify small particle images within the $8$ to $14$ pixel range, particularly at higher particle image densities (Fig.\,\ref{fig:sizeDist}). Therefore, to ensure optimal performance of our models, the defocused particle images should preferably have a minimum diameter of $14$ pixels for $N_s>0.13$, as below this threshold recall may drop significantly. As already discussed, the detection of small particles remains a well-known challenge in computer vision. In future implementations, we plan to address this point by integrating the slicing-aided hyper-inference (SAHI) method \cite{sahi} with YOLOv9.

Ultimately, while our work can be used as an effective first step for three-dimensional defocusing particle tracking, we did not address the determination of the out-of-plane component ($z$). For this task, a similar generalized DNN model that works on any type of optical arrangement is difficult to conceive. Therefore, for the determination of $z$, we refer to existing approaches that rely on experimental calibration procedures. For example, methods based on normalized cross-correlation \cite{Barnkob2021defocustracker} or DNN approaches trained on the specific optical setup \cite{Konig2020} can be used.

Finally, it should be noted that the our models can potentially be applied for the detection of defocused images of non-spherical particles (not tested here), and in combination with recently published methodologies for the determination of the orientation of non-spherical particles \cite{sachs2023spheroids}, cells \cite{sun2025} or micro-organisms \cite{mehdizadeh2024deep}.

\section{Conclusions}\label{sec:Conclusions}

In the present work, we proposed, tested, and demonstrated the capabilities of a generalized deep-learning framework for automatic detection and position estimation of defocused particle images from any kind of optical configuration without compromising uncertainty and recall, referred to as \textit{DefocusTrackerAI}. The final object detection models were based on the YOLOv9-m and YOLOv9-c architectures and trained on an extensive set of synthetic images with the integration of data augmentation techniques.

The comparison with the previous methods using the reference dataset provided in \citep{Barnkob2021}, showed that our models outperform the latest algorithms in terms of spatial resolution, while providing similar uncertainty levels (Fig.\,\ref{fig:NSCritic}). In addition, we demonstrated the generalization to multiple optical configurations using extensive test sets of synthetic images, containing astigmatic and non-astigmatic defocused particle images of varying diameters. Our models were further validated on real DPT experiments using fluorescence, and more importantly, the validation on DPT experiments using shadowgraphy showed that, with the integration of data augmentation techniques, the proposed models can be used beyond conventional DPT applications, including for instance the tracking of sprays and droplets.

In general, we provided a generalized framework that can be applied directly to DPT images of any kind and that will be fully accessible in the public GitLab repository for \textit{DefocusTrackerAI}: \href{https://gitlab.com/goncalo.coutinho/defocustrackerAI-main/-/tree/7e0f11f649ebad50e20dca5b9545f26ca303ebe0/}{DefocusTrackerAI GitLab}. Our approach is expected to be an attractive solution for less experienced users with limited knowledge of DPT, since it is ready to use and does not require additional training or prior knowledge. Note that after successfully recording an experiment, new users simply need to upload the images to Google Colab, open the \textit{DefocusTrackerAI} Jupyter Notebook provided in the GitLab repository, and execute each cell step by step to obtain the 2D coordinates ($x,y$) of the defocused particle images. Future implementations will focus on improving the detection of small particles within the 8 to 14-pixel range with the integration of SAHI. In addition, future tests should include the detection of defocused images of non-spherical particle.

\section*{Acknowledgements}
Gonçalo Coutinho acknowledges the PhD scholarship 2021.04780.BD attributed by Fundação para a Ciência e Tecnologia (FCT). Massimiliano Rossi acknowledges the financial support by the VILLUM foundation under the Grant No. 00036098. Gonçalo Coutinho, Ana Moita a António Moreira acknowledge Fundação para a Ciência e a Tecnologia (FCT) for partially financing the research trough Project Ref. PTDC/EME-TED/7801/2020.

\section*{Data availability statement}
The \textit{DefocusTrackerAI} code used in this article will be fully accessible in the public GitLab repository for \href{https://gitlab.com/goncalo.coutinho/defocustrackerAI-main/-/tree/7e0f11f649ebad50e20dca5b9545f26ca303ebe0/}{DefocusTrackerAI GitLab}. The datasets generated and analysed during the current study are available from the corresponding author upon reasonable request.

\appendix
\section{}

\begin{table}[tph]
\centering
\footnotesize
\setlength{\tabcolsep}{3.5pt} 
 {\fontsize{10}{11}\selectfont 
\begin{tabular}{@{}lll@{}}
\toprule
Parameter             & Faster R-CNN & YOLOv9-m / -c \\ \midrule
Optimizer             & SGDM                  & SGDM                         \\
Epochs                & 20                    & 20 \\
Patience              & 4                     & 4 \\
Minibatch size        & 4                     & 4                            \\
Learning rate         & 1e-4                  & 1e-4                         \\
Learning rate decay   & cosine                & linear                       \\
Warm up period        & 3                     & 3                            \\
Momentum              & 0.9                   & 0.937                        \\
L2 regularization     & 1e-4                  & 1e-4                         \\
Pretrained            & MS-COCO2018           & MS-COCO2018    \\
\bottomrule
\end{tabular}}\caption{Training settings of the object detection models, Faster R-CNN with ResNet50 backbone and FPN, YOLOv9-m and YOLOv9-c with GELAN50 as backbone.}\label{table:trainingsettings}
\end{table}

\begin{table}[tph!]
\setlength{\tabcolsep}{15.5pt} 
\setlength{\arrayrulewidth}{3pt} 
 {\fontsize{10}{11}\selectfont 
\begin{tabular}{@{}lll@{}}
\toprule
Augmentation & Limits          & Probability \\ \midrule
Translate    & [-0.3, 0.3] & 0.5         \\
Scale        & [0.5, 1.5]  & 0.25        \\
Brightness   & [0.7, 1.3]  & 0.2         \\
Contrast     & [0.8, 1.2]  & 0.2         \\
Saturation   & [0.8, 1.2]  & 0.2         \\
Invert       & Not applicable               & 0.2         \\ \bottomrule
\end{tabular}
}\caption{Data augmentation settings for the training of the object detection models, Faster R-CNN with ResNet50 backbone and FPN, YOLOv9-m and YOLOv9-c with GELAN50 as backbone.  } \label{Table:Augmentation}
\end{table}

\bibliography{references}

\end{document}